\documentclass{svproc}
\usepackage{url}

\usepackage[utf8]{inputenc}
\usepackage{amsmath,amssymb,amstext,amsxtra,amsbsy,amsgen}
\usepackage{graphicx}
\usepackage{pgfgantt}
\usepackage{pdfpages}
\usepackage{longtable}
\usepackage{multirow}
\usepackage{makecell}
\usepackage{rotating} 
\usepackage{enumitem}
\usepackage{subcaption}
\renewcommand\appendix{\par
	\setcounter{section}{0}
	\setcounter{subsection}{0}
	\setcounter{figure}{0}
	\setcounter{table}{0}
	\renewcommand\thesection{Appendix \Alph{section}}
	\renewcommand\thefigure{\Alph{section}\arabic{figure}}
	\renewcommand\thetable{\Alph{section}\arabic{table}}
}
\graphicspath{{graphics/}}

\usepackage{xspace}
\usepackage{amsmath,amssymb,amsfonts}

\newcommand{\bs}[1]{\ensuremath{\boldsymbol{#1}}}

\newcommand{\ba}{\ensuremath{\bs a}\xspace}
\newcommand{\bb}{\ensuremath{\bs b}\xspace}
\newcommand{\bc}{\ensuremath{\bs c}\xspace}

\newcommand{\bsf}{\ensuremath{\bs f}\xspace}

\newcommand{\bk}{\ensuremath{\bs k}\xspace}
\newcommand{\bl}{\ensuremath{\bs l}\xspace}

\newcommand{\bp}{\ensuremath{\bs p}\xspace}

\newcommand{\bfs}{\ensuremath{\bs s}\xspace}
\newcommand{\bt}{\ensuremath{\bs t}\xspace}

\newcommand{\bx}{\ensuremath{\bs x}\xspace}
\newcommand{\by}{\ensuremath{\bs y}\xspace}
\newcommand{\bz}{\ensuremath{\bs z}\xspace}

\newcommand{\bH}{\ensuremath{\bs H}\xspace}

\newcommand{\bR}{\ensuremath{\bs R}\xspace}

\newcommand{\bW}{\ensuremath{\bs W}\xspace}
\newcommand{\bX}{\ensuremath{\bs X}\xspace}
\newcommand{\bY}{\ensuremath{\bs Y}\xspace}
\newcommand{\bZ}{\ensuremath{\bs Z}\xspace}

\newcommand{\gm}{\ensuremath{\gamma}\xspace}

\newcommand{\brho}{\ensuremath{\bs \rho}\xspace}

\newcommand{\Del}{\ensuremath{\Delta}\xspace}

\renewcommand{\Re}{\ensuremath{\mathbb{R}}\xspace} 

\newcommand{\SO}{\ensuremath{\mathbb{SO}(3)}\xspace}

\newcommand{\norm}[1]{\ensuremath{\| #1\|}\xspace}

\providecommand{\keywords}[1]
{
	\small	
	\textbf{\textit{Keywords---}} #1
}

\newcommand{\ff}{\ensuremath{\boldsymbol{O}\text{-}\bX\bY\bZ}\xspace}
\newcommand{\mf}{\ensuremath{\boldsymbol{o}\text{-}\bx\by\bz}\xspace}
\newcommand{\bbi}{\ensuremath{\bb_i}\xspace}
\newcommand{\bti}{\ensuremath{\bt_i}\xspace}
\newcommand{\bai}{\ensuremath{\ba_i}\xspace}
\newcommand{\gmb}{\ensuremath{\gamma_\text{f}}\xspace}

\newcommand{\gmt}{\ensuremath{\gamma_\text{m}}\xspace}

\newcommand{\rt}{\ensuremath{r_\text{m}}\xspace}
\newcommand{\rbb}{\ensuremath{r_\text{f}}\xspace}
\newcommand{\rthr}{\ensuremath{\mathbb{R}^3}\xspace}
\newcommand{\ith}{\ensuremath{i}\text{th}\xspace}

\newcommand{\orws}{\ensuremath{\mathcal{W}_\text{O}}\xspace}
\newcommand{\phimax}{\ensuremath{\overline{\phi}}\xspace}
\newcommand{\nso}{\ensuremath{N_\text{S}}\xspace}
\newcommand{\stwo}{\ensuremath{\mathcal{S}_2}\xspace}
\newcommand{\tilrj}{\ensuremath{\widetilde{r}_j}\xspace}
\newcommand{\setwo}{\ensuremath{\mathcal{W}_\text{s}}\xspace}

\newcommand{\srs}{SRSPM\xspace}
\newcommand{\sgp}{SGPM\xspace}
\newcommand{\sps}{6-S\underline{P}S\xspace}
\newcommand{\sfs}{SFS\xspace}
\newcommand{\lsfs}{SFS\xspace}
\newcommand{\dtmon}{SRSPM~1\xspace}
\newcommand{\dtmtw}{SRSPM~2\xspace}
\newcommand{\dtmth}{SRSPM~3\xspace}
\newcommand{\dtmfo}{SRSPM~4\xspace}
\begin{document}
%
\title{A semi-analytical approach for computing the largest singularity-free spheres of a class of $6$-$6$ Stewart-Gough platforms for specified orientation workspaces}
\titlerunning{Singularity-free sphere of a class of $6$-$6$ Stewart-Gough platforms}  
%
\author{Bibekananda Patra, \and Sandipan Bandyopadhyay}
\authorrunning{B. Patra, and S. Bandyopadhyay} 
%
%
\institute{Indian Institute of Technology Madras, Chennai 600036, India,\\
\email{bibeka.patra2@gmail.com, sandipan@iitm.ac.in},\\ 
WWW home page:
\texttt{https://ed.iitm.ac.in/$\sim$sandipan/}
}
\maketitle              

\begin{abstract}
This article presents a method for computing the largest \emph{singularity-free sphere} (\sfs) of a $6$-$6$ Stewart-Gough platform manipulator (\sgp) over a specified orientation workspace. For a fixed orientation of the moving platform, the~\sfs is computed analytically. This process is repeated over a set of samples generated within the orientation workspace, and the smallest among them is designated as the desired~\sfs for the given orientation workspace. Numerical experiments are performed on four distinct architectures of the \sgp to understand their relative performances w.r.t.~\sfs volumes over the same orientation workspace. This study demonstrates the potential utility of the proposed computational method both in analysis and design of SGPMs.

\end{abstract}
%
\section{Introduction}
\label{sc:intro}
\begin{figure}[t]
	\centering
	\begin{minipage}{0.45\textwidth}
		\centering
		\includegraphics[width=0.9\textwidth]{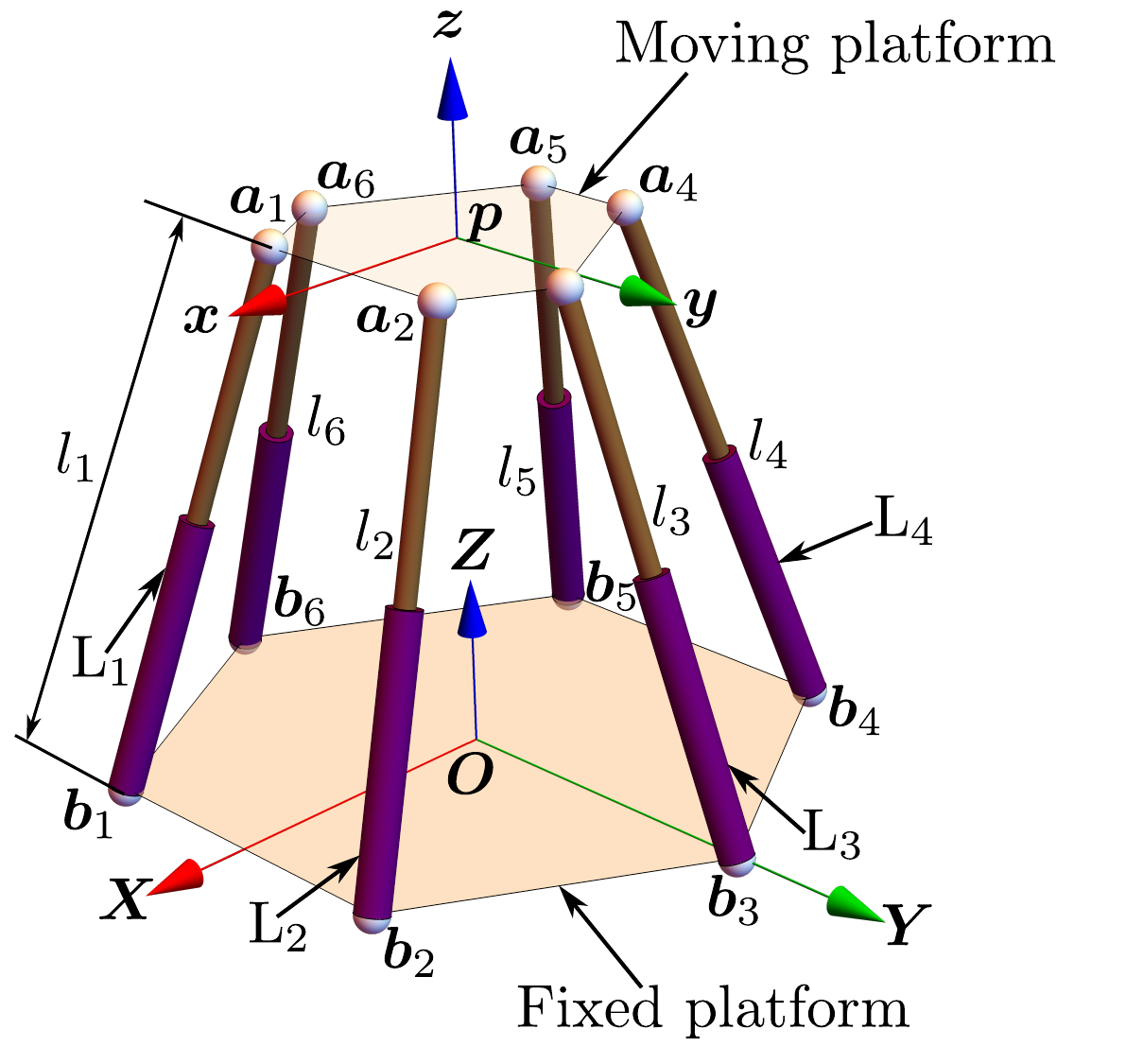}
		\caption{Schematic of a semi-regular Stewart-Gough platform manipulator}
		\label{fg:srspm}
	\end{minipage}
	\hfill
	\begin{minipage}{0.45\textwidth}
		\centering
		\includegraphics[width=\textwidth]{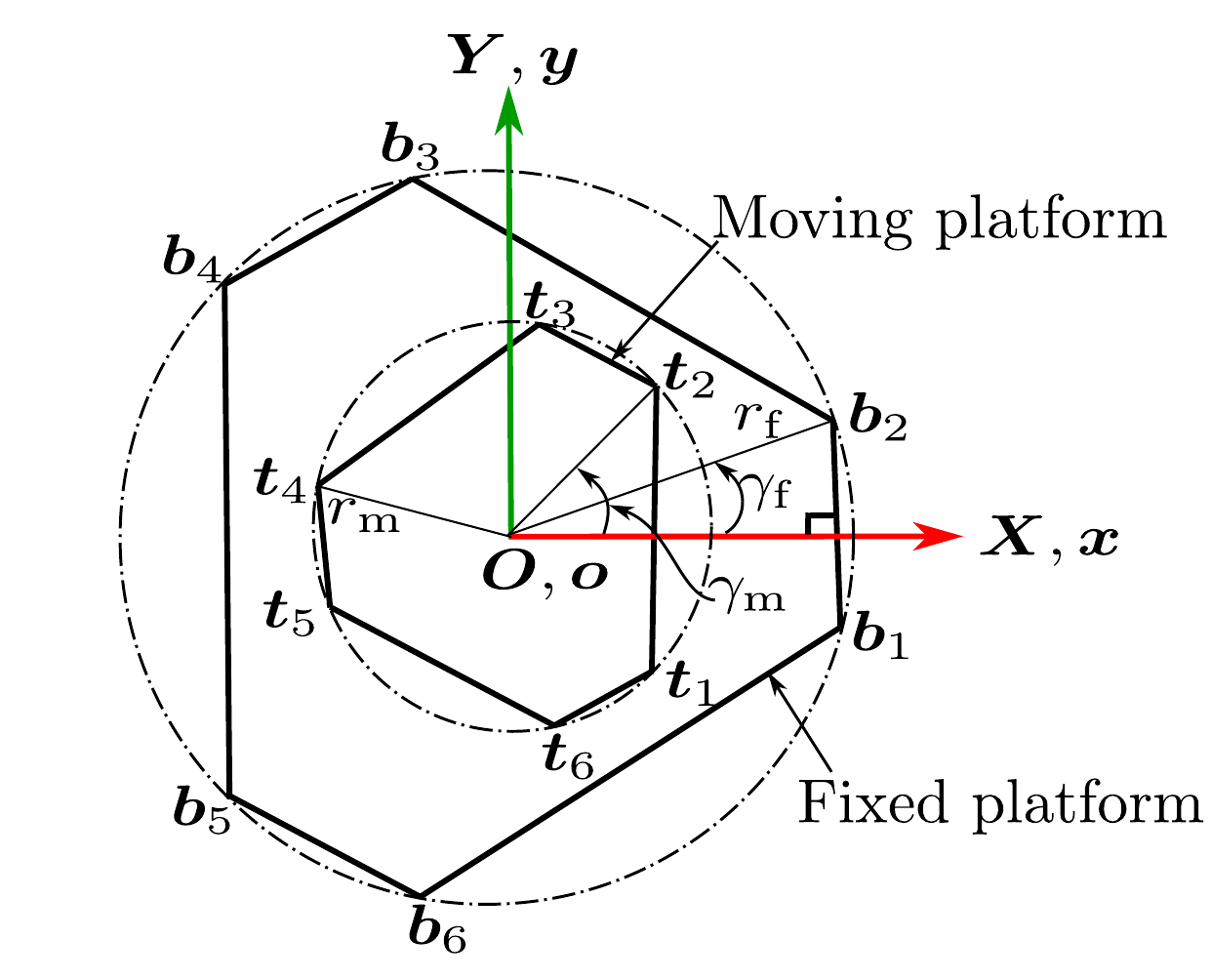}
		\caption{Geometry of the fixed and moving platforms}
		\label{fg:platforms}
	\end{minipage}
\end{figure}
Parallel manipulators are known to have better rigidity, payload capacity, accuracy, and precision, when compared to their serial counterparts of comparable dimensions. However, they have a relatively smaller workspaces due to certain inherent issues, such as the \emph{loss-type singularity}~\cite{GhoslRavani98},~\emph{gain-type singularity}~\cite{Bandyopadhyay2006}, link interference~\cite{Merlet1995}, and limitation of their passive joints' motions~\cite{Ceccarelli2002}. This paper focuses on the identification of a spherical region that is free of gain-type singularities\footnote{The term ``gain-type singularity'' will be referred to as ``singularity'' hereinafter as loss-type singularities
do not occur in Stewart-Gough manipulators.} within the \emph{effective regular workspace}~\cite{PatraS12024} of a class of~$6$-$6$ \emph{Stewart-Gough platform manipulators} (SGPMs)~\cite{Husty1996}, namely, the~\emph{semi-regular Stewart-Gough platform manipulators}\footnote{These have been designated previously as ``semi-regular Stewart platform manipulators'' in the literature; see, e.g.,~\cite{Bandyopadhyay2006,Nag2021}.}~(SRSPMs).
At a singular configuration, the~\srs produces an uncontrollable motion. A common approach to circumvent this issue is to identify contiguous regions within the manipulator's workspace that are free of such singularities. These regions, referred to as the \emph{singularity-free regions} (SFRs) (see, e.g.,~\cite{Pernkopf2006}), are typically chosen to have convex shapes, so as to facilitate their use in path-planning or in the design of such manipulators for specific SFRs.    

The \emph{singularity-free sphere}~(\sfs) of an~\srs has been computed \emph{analytically} in~\rthr for a given orientation as well as in~\SO for a given position of the moving platform (MP) in~\cite{Nag2021}. While these two works use Rodrigues parameters to represent a point in~\SO (corresponding to the orientation of the MP), in~\cite{Li2007}, the~\sfs of the general Gough-Stewart platform manipulator has been obtained using an Euler angle parametrisation. To the best of the authors' knowledge,  a computational methodology for determining the~\sfs of an~\srs for a given orientation workspace (subset of~\SO), represented in terms of~\emph{Rodrigues parameters} in particular, is not available in the existing literature. 

In this article, the \emph{singularity surface} in~\rthr, represented by a \emph{cubic} equation, is obtained using the method described in~\cite{Bandyopadhyay2006}, followed by the analytical computation of the corresponding~\sfs for a constant orientation of the MP as per~\cite{Nag2021}. Since the  \emph{univariate} equation leading to the~\sfs is of degree~24, it is not possible to extend this computation analytically to determine the SFS for a given orientation workspace. Therefore, a ``semi-analytical'' approach is followed for the same, as explained in Section~\ref{sc:sfsorws}: while for a given point in~\SO, the computation is analytical, over a subset of~\SO, it is based on a finite number of samples. In this work,~$299,940$ samples are used, leading to a computation time a little in excess of 9~m, which establishes the utility of the proposed approach. 

The rest of the article is organised as follows. Section~\ref{sc:singusurf} discusses the geometry of the manipulator followed by the derivation of the singularity surface, as well as the computation of the~\sfs for a constant orientation of the MP. The~\sfs for a specified orientation workspace, represented in Rodrigues parameters, is determined in Section~\ref{sc:sfsorws}. Four~\srs architectures are selected to demonstrate the theoretical developments, and their performances are subsequently analysed in Section~\ref{sc:numres}. Finally, the article is concluded in Section~\ref{sc:concl}.
\section{Mathematical preliminaries}
\label{sc:singusurf}
This section describes the geometry of the manipulator, followed by a brief derivation of the singularity condition, and the computation of the~\sfs for a fixed orientation of the MP.
\subsection{Geometry of the SRSPM}
\label{sc:geom}
An~\srs is shown schematically in Fig.~\ref{fg:srspm}.
This manipulator belongs to the \sps class; i.e., it is actuated by six linear actuators connected to the fixed platform (FP) and MP via spherical joints. The lengths of the legs are denoted by~$l_i,i=1,\dots,6$. The centres of the spherical joints are assumed to coincide with the corresponding vertices of the platforms. The geometry of the platforms is depicted in Fig.~\ref{fg:platforms}. The circum-radii of the FP and MP are denoted by~\rbb and~\rt, respectively. Without any loss of generality, the radius~\rbb is considered to be unity, rendering all the linear entities dimensionless.
The origins of the fixed reference frame~\ff and the moving frame~\mf are attached to the centroid of the FP and MP, respectively. The angular displacements of the spherical joints mounted on the FP, measured from the positive~\bX-axis in a CCW manner, are given by:~$\{-\gmb,\gmb,2\pi/3-\gmb, 2\pi/3+\gmb,4\pi/3-\gmb,4\pi/3+\gmb\}$, where~$2\gmb$ is the angle subtended by the two vertices immediately adjacent to the~\bX-axis at the centre of the FP. 
Similarly, the angular displacements of the spherical joints mounted on the MP, measured from the positive~\bx-axis in a CCW manner, are represented by:~$\{-\gmt,\gmt,2\pi/3-\gmt, 2\pi/3+\gmt,4\pi/3-\gmt,4\pi/3+\gmt\}$, where~$2\gmt$ is defined in an analogous manner. The vertices of the FP and MP are denoted by~\bbi and~\bti, respectively, when represented in their respective local frames of reference,~\ff and~\mf (see Fig.~\ref{fg:platforms}).
The origin of the moving frame~\mf corresponds to the point~$\bp=[x,y,z]^\top$ in the fixed frame~\ff.
The orientation of~\mf w.r.t.~\ff is represented by the matrix~$\bR \in \SO$, expressed in terms of \emph{Rodrigues parameters}\footnote{It is well-known that Rodrigues parameters suffer a {\em parametric singularity} at {\em half-turns}. However, this work is not hindered by the same, since~$\phimax < \pi$, as seen in Section~\ref{sc:orwssampl}.}, denoted by~$\bc = [c_1,c_2,c_3]^\top \in~\rthr$ (see~\cite{PatraSO32025} and the references therein). These  parameters are defined as~$\bc = \bk \tan\frac{\phi}{2}$, where~$\bk\in~\mathbb{S}^2$ and~$~\phi \in (0,\pi)$.
The vectors along the legs are defined as:~$\bl_i = \bai - \bbi$, and their lengths are denoted by:~$l_i = \norm{\bl_i}$, where~\bbi and~\bai are the vertices of the FP and MP, respectively, represented in~\ff (see Fig.~\ref{fg:srspm}).
\subsection{Derivation of singularity surface for a fixed orientation of MP}
\label{sc:singudurf}
\begin{figure}[t]
	\centering
	\begin{minipage}{0.35\textwidth}
		\centering
		\includegraphics[width=\textwidth]{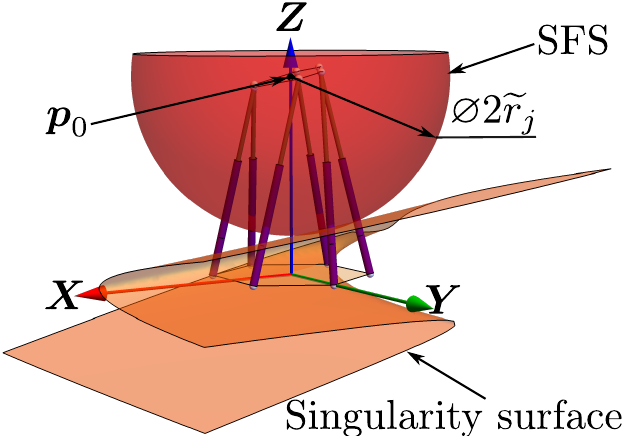}
		\caption{Singularity surface and the corresponding largest~\sfs centred at~$\bp_0=[0,0,2.5000]^\top$ for a given orientation,~$\bc = [0.0639,0.1107,0.2597]^\top$, of the MP of the~\dtmon (see Table~\ref{tb:archdata})}
		\label{fg:surfsfs}
	\end{minipage}
	\hfill
	\begin{minipage}{0.25\textwidth}
		\includegraphics[width=\textwidth]{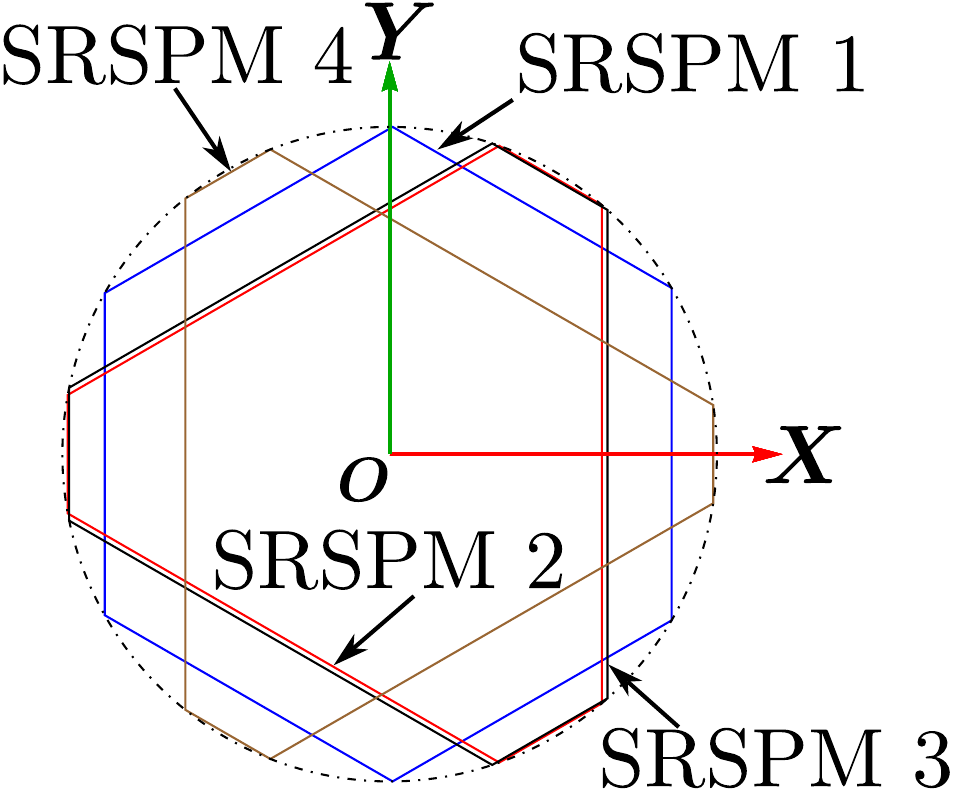}
		\caption{Fixed platforms of the four SRSPMs (see Table~\ref{tb:archdata})}
		\label{fg:fpdtm}
	\end{minipage}
	\hfill
	\begin{minipage}{0.25\textwidth}
		\centering
		\includegraphics[width=\textwidth]{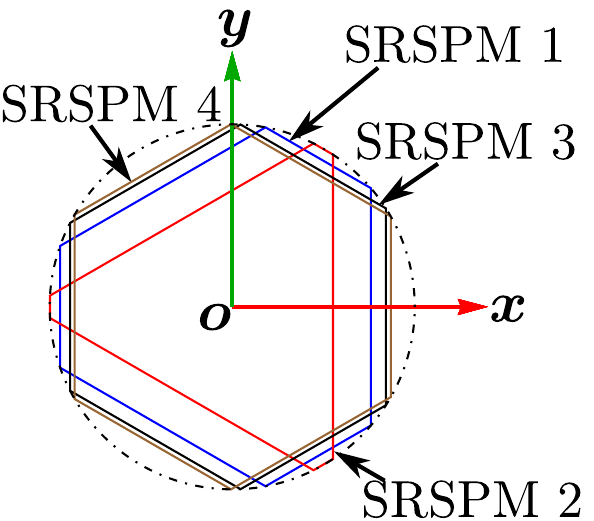}
		\caption{Moving platforms of the four SRSPMs (see Table~\ref{tb:archdata})}
		\label{fg:mpdtm}
	\end{minipage}
\end{figure}
The derivation of the equation of the singularity surface for the~\srs whose~\bX- and~\bx-axes passing through one of the vertices of its FP and MP, respectively, is presented in~\cite{Bandyopadhyay2006}. In this article, the same procedure is followed to obtain the condition of the singularity of the~\srs whose~\bX- and~\bx-axes are along the perpendicular bi-sectors of the line segments joining the vertices~$\bb_1,\bb_2$ of the FP, and the vertices~$\bt_1,\bt_2$ of the MP, respectively, as shown in Fig.~\ref{fg:platforms}. This has been explained below briefly for the sake of completeness.

The loop-closure equations of the manipulator are written in terms of the unit leg-vector~$\bfs_i$, as follows:~$\bbi + l_i\bfs_i = \bp + \bR\bti, \ i = 1,\dots,6$.
The vector~$\bfs_i$ is obtained as:~$\bfs_i = (\bp + \bR \bti - \bbi)/l_i$.
The \emph{wrench} along the~\ith leg,~$\bW_i$, resulting from the force~$f_i$ acting along that leg, is expressed as:~$\bW_i = \left(f_i\bfs_i; (\bR\bti) \times f_i\bfs_i\right)$. The net wrench acting on the MP is given by~$\bW = \sum_{i=1}^{6} \bW_i$, and it can be written as:~$\bW = \bH \bsf$, where~$\bH \in \Re^{6\times6}$ is known as the~\emph{wrench transmission matrix}, while~$\bsf = [f_1,\dots,f_6]^\top  \in \Re^6$. The~\ith column of \bH is found as:~$\frac{1}{l_i}\left[\bp + \bR\bti - \bbi; (\bR \bti) \times (\bp-\bbi)\right] \in \Re^6$.

The singularity of the manipulator leads to the degeneracy of the column space of~\bH. Therefore, the condition for singularity may be derived as:
\begin{align}
	\det(\bH) = 0. \label{eq:detH}
\end{align}
At such a singularity, even finite wrenches in the left-nullspace of~\bH cannot be supported by \emph{any} finite~$\bsf$.
 
Equation~(\ref{eq:detH}) describes the singularity surface in terms of task-space coordinates $\{x,y,z,c_1,c_2,c_3\}$, as well as the architecture parameters of the manipulator,~$\brho = [\rt,\gmb,\gmt]^\top$. Following the simplification schemes introduced in~\cite{Bandyopadhyay2006}, it can be cast in the closed-form in terms of~$\{x,y,z\}$ as:
\begin{alignat}{3}
 &e(x,y,z)&&:= &&\frac{\left(3\rt c_0\right)^3 \sin\gm}{l_1l_2l_3l_4l_5l_6} g(x,y,z)= 0,~\text{where}~c_0 = 1+c_1^2 +c_2^2 + c_3^2,~\gm = \gmb-\gmt,~\text{and}  \nonumber\\
&g (x,y,z)&&:= &&
a_0 x^2 z + a_1 x^2 + a_2 x y z + a_3 x y +  
a_4 x z^2 +a_5 x z + a_6 x + a_7 y^2 z + a_8 y^2 \nonumber\\ & && &&+a_9 y z^2 + a_{10} y z + a_{11} y + 
a_{12} z^3 + a_{13} z^2 + a_{14}z +a_{15}.
\label{eq:singsurf2}
\end{alignat}
The coefficients\footnote{The expressions of the coefficients of~$g(x,y,z)$ are available in the closed form, but not presented here due to limitations in space.}~$a_0, \dots, a_{15}$ of~$g(x,y,z)$ are functions of the orientation parameters~\bc and the architecture parameters~\brho. It is obvious that in a physical manipulator~$l_i \ne 0$, and~$\rt  \ne  0$. Additionally,~$c_0=1+\bc\cdot\bc \ne 0~\forall~\bc \in \rthr$. However, if~$\sin \gamma = 0$, then~$\gmb = \gmt$ or~$|\gmb-\gmt| = \pi$. The former condition is known as an \emph{architecture singularity}~\cite{MaSing1991} of an SRSPM, whereas the latter is infeasible since~$\gmt,\gmb \in \left(0,\frac{2 \pi}{3}\right)$. For~$\sin\gm \ne 0$, the singularity condition reduces to~$g(x,y,z) = 0$. The corresponding surface in~\rthr is shown in Fig.~\ref{fg:surfsfs}.
\subsection{The singularity-free sphere for a fixed orientation of the MP}
\label{sc:sfsfixed}
The~\sfs for a fixed orientation of the MP is computed analytically using \emph{Sylvester's dialytic} method in~\cite{Nag2021} via the derivation of a 24-degree univariate polynomial in one of the variables~$\{x,y,z\}$, which is subsequently solved using the~\emph{generalised eigenproblem} approach. The details are omitted here for want of space. 
The largest~\sfs is tangent to the singularity surface described by~$g(x,y,z) = 0$, without intersecting it at any other point, as shown in Fig.~\ref{fg:surfsfs}. The centre of the~\sfs is chosen to be the {\em neutral position}~$\bp_0 = [0,0,z_0]^\top$ of the manipulator, where~$z_0$ is its \emph{neutral height}\footnote{The neutral height, introduced in~\cite{Nabavi2018}, locates a point on the~\bZ-axis of the base frame of reference, which, in turn, serves as the position component of the {\em home pose} of the manipulator.}.
\section{Computation of SFS for a given orientation workspace}
\label{sc:sfsorws}
In the above, the~\sfs is computed for a given orientation of the MP.
This section presents an extension of the same over a given orientation workspace, denoted by~$\orws \subset \SO$, which is achieved  via the following steps.
\begin{enumerate}
	\item First, the orientation workspace~$\orws$ is described as a sphere (following~\cite{PatraSO32025}). It is subsequently sampled to generate a discrete set, denoted by~\setwo,  of cardinality~\nso. 
	\item Next, for each sample in~\setwo, i.e., a distinct orientation of the MP, the largest~\sfs is computed analytically.
	\item Finally, the smallest among the~\nso SFSs obtained in Step~2, is identified as the~\sfs over~\setwo (and subject to the resolution of the sampling process of~\orws). The radius of~\stwo is denoted by~$r_2$.
\end{enumerate}
The above steps are elaborated further in the following. 
\subsection{Description of the orientation workspace and its discretisation}
\label{sc:orwssampl}
The detailed descriptions of the orientation workspace~\orws as a sphere may be found in~\cite{PatraSO32025}. As mentioned in Section~\ref{sc:geom}, the matrix~\bR is represented in terms of Rodrigues parameters,~$\bc = \bk\tan(\phi/2)$, where~\bk represents the {\em axis of rotation} and~$\phi$ the corresponding angle, associated with the given~\bR. In this framework,~\orws is described as a sphere of radius~$\tan(\phimax/2)$, i.e., it is a subset of~\SO consisting of {\em all possible rotations~$\phi \in \left(0,\phimax \right]$, about all~$\bk \in \mathbb{S}^2$}. Based on this interpretation, various schemes for sampling~\orws were compared in~\cite{Patra2023}, among which \emph{uniform sampling based on regular placement}~(USRP) emerged as the best choice in terms of performance\footnote{The schemes presented in~\cite{Patra2023} sample~$\mathbb{S}^2$ and~$\Re$, i.e., the domains of~\bk and~$\phi$, {\em separately}, and as such, it is obvious that these {\em do not} generate {\em uniform} samples in~\SO. Nevertheless, these schemes allow reliable and controlled sampling of~\SO that leads to physically intuitive results, as demonstrated in this work.}.

The following section discusses the computation of the~\sfs for a given set of samples in~\orws, as well as identification of the~\sfs~\stwo for a specified~\orws.
\subsection{Computation of largest SFS for given orientation workspace}
\label{sc:sfsconst}
\begin{table}[t]
	\centering
	\begin{minipage}{\textwidth}
		\centering
		\caption{Architecture parameters of FP and MP  of four SRSPMs (see Figs.~\ref{fg:fpdtm},~\ref{fg:mpdtm})}
		\label{tb:archdata}
		\begin{tabular}{|l|l|r|r|r|r|}
			\hline
			\makecell{Significance} & Symbol & \dtmon & \dtmtw & \dtmth& \dtmfo\\
			\hline 
			\makecell[l]{Half of the angle subtended by~$\bb_1$ \\and~$\bb_2$ at the centre of the FP} & \gmb &0.5328 & 0.8652 & 0.8425 & 0.1506\\
			\hline
			\makecell[l]{Half of the angle subtended by~$\bt_1$ \\and~$\bt_2$ at the centre of the MP} & \gmt &0.7073 & 0.9875 & 0.5694 & 0.5173\\
			\hline
		\end{tabular}
	\end{minipage}
\end{table}
The largest~\sfs for a given~\orws is identified through the following steps.
\begin{enumerate}
	\item For the set~\setwo, the feasibility of the neutral position~$\bp_0 = [0,0,z_0]^\top$ is checked by verifying if it satisfies~$g(0,0,z_0) = a_{12}z_0^3 + a_{13}z_0^2 + a_{14}z_0 + a_{15} = 0$. If so,~$\bp_0$ lies on the singularity surface and~$r_2 = 0$; otherwise, it is said to be safe, and the next step in computing the~\sfs~\stwo is pursued. 
	\item For each sample in~\setwo, i.e., a given orientation of the~MP, the radius of the~\sfs (denoted by~\tilrj) is computed (see Section~\ref{sc:sfsfixed}). Finally, the radius of the~\lsfs~\stwo is determined as~$r_2 = \min(\widetilde{r}_1,\dots, \widetilde{r}_{N_\text{s}}).$
\end{enumerate}
\section{Numerical results}
\label{sc:numres}
\begin{table}[t]
	\centering
	\begin{minipage}{0.45\textwidth}
			\centering
		\caption{Controlling parameters associated with discretisation of~\orws}
		\label{tb:orwsparam}
		\begin{tabular}{|l|r|}
			\hline
			Parameter &\makecell{ Range/value}\\
			\hline
			$\phi$ & $[1^\circ,30^\circ]$\\
			\hline
			$\Del \phi$ & $1^\circ$\\
			\hline
			\nso & $299,940$\\
			\hline
		\end{tabular}
		\centering
		\caption{Values of~$r_2$ and execution times for four SRSPMs over~\orws}
		\label{tb:r2dtm}
		\begin{tabular}{|c|r|r|}
			\hline
			\srs & \makecell{$r_2$} & \makecell{Execution time} \\
			\hline
			\dtmon  & 1.1560 & 9~m~17~s\\
			\hline
			\dtmtw & 1.4466 & 9~m~11~s\\
			\hline
			\dtmth & 1.6039 & 9~m~23~s\\
			\hline
			\dtmfo & 1.7356 & 9~m~12~s\\
			\hline
		\end{tabular}
	\end{minipage}
	\hfill
	\begin{minipage}{0.45\textwidth}
		\centering
		\includegraphics[width=\textwidth]{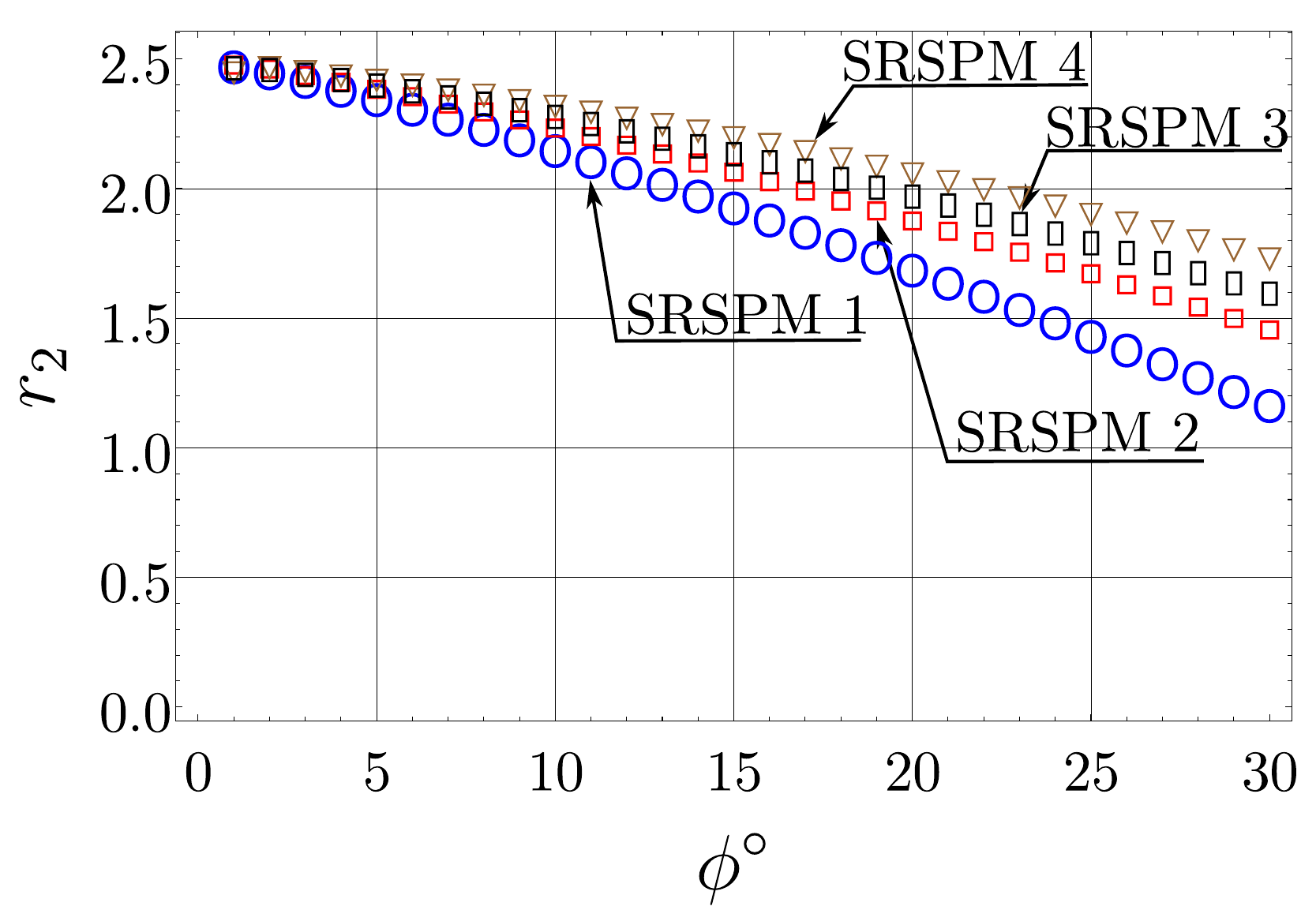}
		\captionof{figure}{Variations of~$r_2$ w.r.t.~$\phi \in [1^\circ, 30^\circ]$ corresponding to four SRSPMs}
		\label{fg:varr2}
	\end{minipage}
\end{table}
The purpose of the computational framework described above is to provide a means to compute a single scalar measure of performance that can be used to compare different designs of SRSPM from a certain kinematic perspective. Towards that, the theoretical developments presented above are demonstrated via applications to four architecturally distinct SRSPMs. The circum-radii of the MP of the four SRSPMs are chosen to be identical, i.e.,~$\rt=0.5$, while the angles~\gmb and~\gmt of~\dtmon,~$\dots,$~\dtmfo (listed in Table~\ref{tb:archdata}) are chosen randomly. The corresponding FPs and MPs are shown in Figs.~\ref{fg:fpdtm},~\ref{fg:mpdtm}, respectively.
All angles in this section are in radians unless stated otherwise.
To focus only on the effect of the platform geometries, the neutral height is fixed at~$z_0 = 2.5$ for all the SRSPMs in this study. The parameters associated with the sampling of~\orws are listed in Table~\ref{tb:orwsparam}. 
All computations are performed using a \textsf{C++} programme employing a single core of an AMD\textsuperscript{\textregistered} Ryzen\textsuperscript{TM} 9 7950X CPU running at 4.5~GHz. All numbers are rounded up to four digits after the decimal point.

For a randomly chosen sample in~\setwo, i.e.,~$\bc = [0.0639,0.1107,0.2597]^\top$, 
the singularity surface,~$g(x,y,z)=0$, (see Fig.~\ref{fg:surfsfs}) is derived for the~\dtmon architecture mentioned in Table~\ref{tb:archdata}.
The radius of the~\sfs centred at~$\bp_0$ is found as:~$\tilrj = 1.9864$, which is tangent to the above surface. Table~\ref{tb:r2dtm} lists the radius~$r_2$ of the~\sfs corresponding to the the four designs of the SRSPM, as well as the  execution times, for~$299,940$ samples in~\setwo.
Figure~\ref{fg:varr2} shows the variations of~$r_2$ w.r.t.~$\phi$ for the four manipulators under the study. As expected,~$r_2$ decreases monotonically with~$\phi$, but the rates are decided by the architecture parameters, which motivates further studies, e.g., a quest to determine the optimal architecture parameters for the best performance w.r.t.~$r_2$. Nevertheless, it is important to note here that such a study is made possible by the proposed computational framework, in which the extents of the orientation workspace is captured by a single scalar, thus affording an objective basis for comparing various designs. 
\section{Conclusions}
\label{sc:concl}
This paper presents a semi-analytical computational framework to compute the largest~\sfs of the~\srs for a given orientation workspace, and to quantitatively compare various designs from this perspective. The orientation workspace is discretised into a finite set of samples, and the smallest the~\sfs is computed analytically over this set, to identify the~\sfs for the given workspace. The method is demonstrated via application to four manipulators of the~\srs architectures. The proposed method is computationally inexpensive, and effective in providing a means for analysis of different designs, as well as an objective comparison among them. It is hoped to be useful in the design as well as path planning of such manipulators.
\bibliographystyle{spmpsci} 
\bibliography{Bibek_sfs}
\end{document}